\documentclass[runningheads]{llncs}
\usepackage[T1]{fontenc}
\usepackage{graphicx}
\usepackage[toc,page]{appendix}

\usepackage{graphicx}
\usepackage{latexsym}

\usepackage{amsmath, amssymb, nccmath}
\usepackage{booktabs}

\usepackage{subcaption}

\DeclareMathOperator*{\argmin}{arg\,min}
\DeclareMathOperator{\EX}{\mathbb{E}}

\begin{document}
\title{Learning Solutions of Stochastic Optimization Problems with Bayesian Neural Networks}
%
%
\author{Alan A. Lahoud \and
	Erik Schaffernicht \and
	Johannes A. Stork}
%
%
\institute{Center for Applied Autonomous Sensor Systems (AASS)\\
	Örebro University, Örebro, Sweden}
%
%
%
\maketitle              
\begin{abstract}
Mathematical solvers use parametrized Optimization Problems (OPs) as inputs to yield optimal decisions.
In many real-world settings, some of these parameters are unknown or uncertain.  
Recent research focuses on predicting the value of these unknown parameters using available contextual features, aiming to decrease decision \emph{regret} by adopting end-to-end learning approaches. 
However, these approaches disregard prediction uncertainty and therefore make the mathematical solver susceptible to provide erroneous decisions in case of low-confidence predictions.
We propose a novel framework that models prediction uncertainty with Bayesian Neural Networks (BNNs) and propagates this uncertainty into the mathematical solver with a Stochastic Programming technique.
The differentiable nature of BNNs and differentiable mathematical solvers allow for two different learning approaches: In the \emph{Decoupled} learning approach, we update the BNN weights to increase the quality of the predictions' distribution of the OP parameters, while in the \emph{Combined} learning approach, we update the weights aiming to directly minimize the expected OP's cost function in a stochastic end-to-end fashion. We do an extensive evaluation using synthetic data with various noise properties and a real dataset, showing that decisions \emph{regret} are generally lower (better) with both proposed methods.
\end{abstract}
\begin{figure}
	\includegraphics[width=\textwidth]{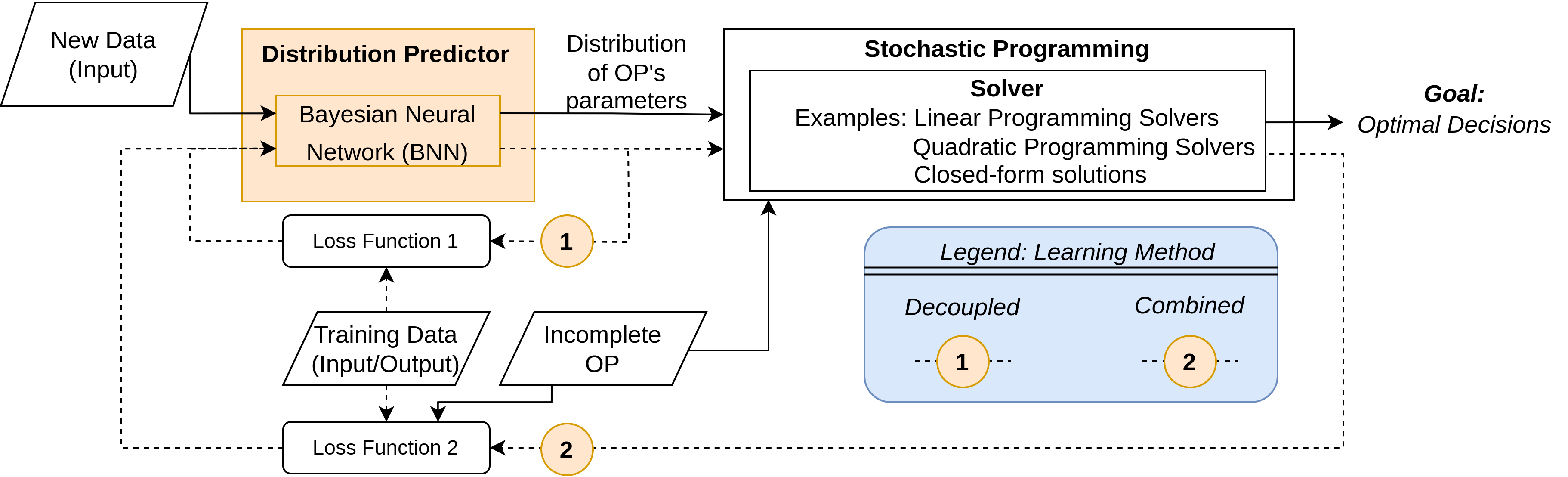}
	\caption{
		A diagram depicting the proposed addition of a BNN distribution predictor block before solving data-driven OPs with a mathematical solver in a stochastic fashion. The solid lines illustrate the decision inference given new input data, while the dotted lines indicate the learning process, where the BNN weight updates can be computed based on prediction quality (\emph{Decoupled} learning, 1) or decision quality (\emph{Combined} learning, 2).} \label{fig:contributions}
	
\end{figure}%

\section{Introduction}

A mathematical solver uses optimization techniques to find solutions for Optimization Problems (OPs), defined by a cost function and a feasible set, aiming to reach global optima. This process minimizes the cost function by finding the best decisions within the feasible set. In real-world scenarios, some parameters of the OP might be unknown during decision-making and must be estimated using available information. Examples include optimizing trading decisions (e.g., minimizing risk) based on unknown market fluctuations; and optimizing energy scheduling considering estimated energy demand. This paper focuses on modeling the uncertainty of these unknown parameters (e.g, market fluctuations; energy demand) of the OP using input-output training data, where the target variables represent the unknown OP parameters.

From a Machine Learning perspective, one could predict those unknown parameters in a supervised fashion using contextual features. Predicting these parameters completes the OP (albeit with estimated parameters), allowing mathematical solvers to seek optimal decisions. However, inaccuracies in predictions due to noise or lack of data can affect decision quality in ways that may vary significantly depending on the OP type \cite{Ifrim2012tradapp,GRIMES2014276}. Some recent methods align the loss function with the OP's cost function to avoid this problem \cite{wilder2018melding,elmacht2017spo,mandi2020interior}. However, they remain deterministic and thus ignore model and data uncertainty. 

In contrast, Stochastic Programming approaches provide a more robust solution by considering different sets of possible OP parameters, modeling them as distributions. Mathematical solvers then find optimal decisions that account for this distribution \cite{birge2011introduction}. In this paper, we explore the treatment of data-driven OPs as Stochastic OPs, introducing a novel approach to model the distribution of their unknown parameters focusing on decision quality. To achieve this, we leverage Bayesian Neural Networks' (BNNs) architecture due to its significant advantages in modeling complex and non-linear data relationships using gradient descent approaches. 

The main motivation of our work is to enhance decision quality by connecting probabilistic models to data-dependent OPs within a task-oriented learning framework. More specifically, our contributions include:

\begin{itemize}
	\item Our first method, the \emph{Decoupled} Learning, integrates existing learning techniques within BNNs to predict the actual distribution of the unknown parameters of conditional Stochastic OPs. Then, it propagates the predicted distribution to provide decisions that minimizes the expected final task cost in a Stochastic Programming fashion (See Figure \ref{fig:contributions}, \emph{Decoupled}).
	
	\item Our second method, the \emph{Combined} Learning, introduces an end-to-end gradient computation. It innovates the way of learning BNNs by implementing a task-oriented method, where the goal is not to fit the data distribution, but to learn a distribution that directly minimizes the OP cost in a Stochastic Programming fashion (See Figure \ref{fig:contributions}, \emph{Combined}). 
	
	\item Through experimental results, we show that our \emph{Decoupled Learning} method generally outperforms existing uncertainty estimation methods, while our \emph{Combined Learning} method surpasses previous end-to-end learning approaches. We further delineate the specific scenarios where either the \emph{Combined} or the \emph{Decoupled} approach holds an advantage.
	
\end{itemize}

\section{Background}

In this section, we review the notation on data-driven OPs and BNNs since we use them as our probabilistic model.

\subsection{Problem Formulation}

\paragraph{OPs under uncertainty.} 
In OPs under uncertainty, the goal is to identify the optimal decisions $\boldsymbol{z}^* \in S \subset \mathbb{R}^{d_z}$, which depend on an unknown parameter $\boldsymbol{y} \in \mathbb{R}^{d_y}$. This is expressed as $\boldsymbol{z}^*({\boldsymbol{y}}) = \argmin_{\boldsymbol{z}} f(\boldsymbol{z}, \boldsymbol{y})$ subject to $\boldsymbol{z} \in {S}$, where $f$ is a cost function (task loss) and $S$ is the feasible decision set. Given $\boldsymbol{y}$'s uncertainty, a common approach is to approximate it with a parametric distribution $\hat{\boldsymbol{y}}$ \cite{birge2011introduction,hannah2015stochastic,bayraksan2015data}, leading to the approximation $\boldsymbol{z}^*({\boldsymbol{y}}) \approx \boldsymbol{z}^*(\hat{\boldsymbol{y}})$ where $\boldsymbol{z}^*(\hat{\boldsymbol{y}}) := \argmin_{\boldsymbol{z}} \EX_{\hat{\boldsymbol{y}}} [f(\boldsymbol{z}, \hat{\boldsymbol{y}})]$ subject to $\boldsymbol{z} \in {S}$. Here, $\hat{\boldsymbol{y}}$ is a predicted distribution. This equation captures many real-world problems related to decision-making under uncertainty \cite{powell2019unified}, where practitioners need to make decisions before having the actual observation of the outcome and aim to minimize the cost $f$ in the long term on average. With specific combinations of OPs (i.e., $f$ and $S$) and simple parametric distributions for $\hat{\boldsymbol{y}}$, the argmin can be reduced in a way that standard mathematical solvers (e.g., linear programming solvers, quadratic programming solvers) are able to provide optimal decisions \cite{birge2011introduction}. By varying the form of $f$ and $S$, it is also possible to represent a broad set of known problems such as conditional values at risk and chance-constrained problems \cite{bayraksan2015data}. In general, we are interested in OP formulations where propagating uncertainty from predictions to the OP is effective. This is valid when the equality $\argmin_{\boldsymbol{z}} \EX_{\hat{\boldsymbol{y}}} [f(\boldsymbol{z}, \hat{\boldsymbol{y}})]$ = $ \argmin_ {\boldsymbol{z}} f(\boldsymbol{z}, \EX{[\hat{\boldsymbol{y}}]})$ does not hold, otherwise solving the expectation in the prediction step would lead to the same result as solving the expectation in the decision step. Appendix A provides details regarding this limitation.

\paragraph{Dataset and Problem Statement.}
Let us denote the input-output training data (let superscript $t$ refer to training samples) as $\mathcal{D}^t=\{ ( \boldsymbol{x_i}^t, \boldsymbol{y_i}^t ) \}_{i=1}^{N^t}$. This paper focus on learning a probabilistic model $h_{\omega}$ that outputs predicted distributions for the unknown parameters, i.e., $h_{\omega} := P_{\omega}(\boldsymbol{y} \mid \boldsymbol{x})$. In our first proposed method (\emph{Decoupled}), we explore the case where data-driven decisions are achieved by trying to approximate the model output to the actual unknown parameters distributions $P(\boldsymbol{y} \mid \boldsymbol{x})$, while in our second proposed method (\emph{Combined}), we consider a direct minimization to the expected loss of the downstream task, which although can lead to considerable differences between $P(\boldsymbol{y} \mid \boldsymbol{x})$ and $P_{\omega}(\boldsymbol{y} \mid \boldsymbol{x})$, the final goal can still be achieved. In both cases, data-driven decisions  $\boldsymbol{z}^*({\boldsymbol{x}, h_{\omega}})$ are provided as	
\begin{equation}
\label{eqn:PaO2}
\begin{aligned}
\boldsymbol{z}^*({\boldsymbol{y}}) \approx \boldsymbol{z}^*({\boldsymbol{x}, h_{\omega}}) := &\argmin_{\boldsymbol{z}} \EX_{{\hat{\boldsymbol{y}} \sim h_{\omega}(\boldsymbol{x})}} [f\big(\boldsymbol{z}, \hat{\boldsymbol{y}}\big)] \\
&\text{subject to } \boldsymbol{z} \in {S}
\end{aligned}
\end{equation}
where $\boldsymbol{x} \in \mathbb{R}^{d_x}$ is a feature vector conditioning the unknown parameters $\boldsymbol{y}$ and is available at decision-making time.
We restrict the problem definition to scenarios where decisions do not impact the actual observations of $\boldsymbol{y}$, a common assumption also reflected in prior works \cite{elmacht2017spo,donti2019mloc}.

\subsection{Probabilistic Model as Bayesian Neural Networks}

We have selected BNNs to represent the probabilistic model $h_{\omega}$ due to the following key characteristics. First, BNNs are adept at modeling uncertainty, both epistemic and aleatoric \cite{kendall2017uncertainties,hullermeier2021aleatoric}. Second, the flexibility of BNNs allows them to capture complex relationships between inputs and outputs, making them suitable to be applied into a wide range of datasets. Finally, BNNs are compatible with gradient descent optimization methods, an attribute that is particularly valuable in our proposed \emph{Combined} method due to its end-to-end learning manner. 

In short, BNNs are Neural Networks that contain stochastic components \cite{blundell2015weight,jospin2022hands}. In Variational Inference, BNNs' weights are treated as a distribution $Q$ parametrized by $\theta$ (e.g., Gaussian class), and the aim is to optimize the evidence lower bound (ELBO). The ELBO optimization is often rewritten as
\begin{equation}
\label{eqn:ELBO}
\begin{aligned}
\theta^* = \argmin_{\theta} \EX_{\omega \sim Q_{\theta}(\omega)} [C_{\omega} - \sum_{i=1}^{N^t} \log P(\boldsymbol{y_i}^t \mid \omega, \boldsymbol{x_i}^t)]
\end{aligned}
\end{equation}
where $C_{\omega} = \log Q_{\theta}(\omega) - \log P(\omega)$ works as a regularization term between the weights' distribution and their provided prior distribution $P(\omega)$. Also, assuming a Gaussian likelihood and $\hat{\boldsymbol{y}}$ continuous, and following \cite{kendall2017uncertainties}, the negative log-likelihood term of the above equation is proportional to a data loss that captures epistemic and aleatoric uncertainty, and can be written as \\ $\frac{1}{N^t} \sum_{i=1}^{N^t}\exp(-{h^{\sigma}_{\omega}(\boldsymbol{x_i}^t)}) (\boldsymbol{y_i}^t-h^{\mu}_{\omega}(\boldsymbol{x_i}^t))^2 + h^{\sigma}_{\omega}(\boldsymbol{x_i}^t)$,
where both $h^{\mu}_{\omega}(\boldsymbol{x_i}^t)$ and $h^{\sigma}_{\omega}(\boldsymbol{x_i}^t)$ are BNN outputs representing a stochastic mean and a stochastic variance of the predicted distribution. In practice, $M^t$ weights combinations are sampled to approximate Equation \eqref{eqn:ELBO} \cite{blundell2015weight}, and backpropagation is used to compute gradients with the help of the reparametrization trick \cite{kingma2013auto}. Once $\theta$ is trained, predictions are sampled from the BNN as $h_{\omega}(\boldsymbol{x_i}) =  h^{\mu}_{\omega}(\boldsymbol{x_i}) + \epsilon \circ \sqrt{\exp(h^{\sigma}_{\omega}(\boldsymbol{x_i}))}$ where $\omega \sim Q_{\theta}({\omega})$, $\epsilon \in \mathbb{R}^{d_y}$ is a sample from the multivariate normal distribution and $\circ$ denotes element-wise multiplication.

\section{Methods}

This section presents two methods to learn a BNN $h_{\omega}$. The \emph{Decoupled} Learning focus on approximating the Stochastic OPs' parameters to their actual distribution, while the \emph{Combined} Learning learns a distribution for the Stochastic OPs' parameters by minimizing the OP cost function directly. In both cases, new decisions are made by propagating learned Stochastic OPs' parameters to solve Equation \eqref{eqn:PaO2} for a new set of input data.

\subsection{Decoupled Learning with BNN}
\label{methods1}

If a trained BNN $h_{\omega}$ fits the actual data accurately, then $\boldsymbol{z}^*(h_{\omega}, \boldsymbol{x}) = \boldsymbol{z}^*(\boldsymbol{y})$, indicating that the model leads to optimal data-driven decisions. This insight serves as a motivation for our \emph{Decoupled} framework. This approach initially leverages common BNN learning techniques to approximate the data distribution of the OP parameters directly from the training data $\mathcal{D}^t$. Then, it integrates prediction samples into a Stochastic Programming block.

Inferring decisions from a trained BNN is not straightforward because Equation \eqref{eqn:PaO2} needs to be solved. More specifically, the expected value operator of the downstream task cost function in this equation makes it to be non-trivial. Therefore, we approximate the expectation by sampling $M$ predictions from the learned model, denoted as $\hat{\boldsymbol{y}}^{(j)} \sim h_{\omega}(\boldsymbol{x})$, and then we propagate those predictions into the argmin operator of a single and complete Stochastic OP with no unknown parameters as follows:
\begin{equation}
\label{eqn:PaOMonteCarlo}
\begin{aligned}
\boldsymbol{z}^*(\boldsymbol{x},h_{\omega}, M) = &\argmin_{\boldsymbol{z}} \frac{1}{M}\sum_{j=1}^{M}f\big(\boldsymbol{z}, \hat{\boldsymbol{y}}^{(j)}\big) \qquad \\
&\text{subject to } \boldsymbol{z} \in {S}.
\end{aligned}
\end{equation}

In order to understand better the scalability of this complete Stochastic OP, the cost function $f$ can be rewritten as $f_d(\boldsymbol{z_d}) + \frac{1}{M} \sum_{j=1}^{M} f_s(\boldsymbol{u}^{(j)})$ (as done in \cite{rockafellar2000optimization,rudin2018newsvendor}). Here, $f_d$ represents a deterministic objective function with $\boldsymbol{z_d} \in \mathbb{R}^{d_{z_d}}$ as decision variables, and $f_s$ is a stochastic objective function with $\boldsymbol{u} \in \mathbb{R}^{d_{u}}$ as auxiliary decision variables that depends on prediction samples. The transformation into a Stochastic OP makes the number of decision variables to increase from $d_{z}$ to $d_{z_d} + M*d_{u}$. This transformation and the relationship between the values of $d_{z}$, $d_{z_d}$ and $d_{u}$ depends on the OP's specific structure, which we detail for our experimental problems in Appendices B and C. It is noteworthy that $M$ works as a hyperparameter. A higher value of $M$ provides a more representative estimate of the unknown OP parameters' distribution, resulting in a better decision result. However, it also increases the size of decision variables in the Stochastic OP, making it more time consuming to solve. Careful choice of $M$ is therefore essential to find a balance between solution accuracy and computational efficiency.

\subsection{Combined Learning with BNN}
\label{methods2}

Leveraging the availability of the OP information during training, i.e., $f$ and $S$ are given, decisions can be computed and evaluated during training time. This context drives us to refine the BNN training process, transforming it into a \emph{Combined} learning-optimization differentiable block in order to directly enhance the decision quality of the downstream task. To achieve this, we leverage the same BNN structure as the previous method, but we modify the loss function to minimize the OP cost $f$ subject to $S$. The forward process of the BNN involves computing $M^t$ output samples from the training input as $h_{\omega}(\boldsymbol{x}^t) =  h^{\mu}_{\omega}(\boldsymbol{x}^t) + \epsilon \circ \sqrt{\exp(h^{\sigma}_{\omega}(\boldsymbol{x}^t))}$. Then, the argmin value $\boldsymbol{z}^*({\boldsymbol{x}^t, h_{\omega}, M^t})$ (decision) is calculated following Equation \eqref{eqn:PaOMonteCarlo} and evaluated within the OP cost function $f$.  Additionally, we introduce a regularization term $C_{\omega}$ to the end-to-end loss to address overfitting and convergence difficulties. The combined loss function is then expressed as
\begin{equation}
\label{eqn:ENDLOSS}
\begin{aligned}
\theta^* = &\argmin_{\theta} \big( \EX_{\omega}[C_{\omega}]  + \frac{K}{N^t} \sum_{i=1}^{N^t}
f\big(\boldsymbol{z}^*(h_{\omega}, \boldsymbol{x_i}^t, M^t), \boldsymbol{y_i}^t\big) \big)
\end{aligned}
\end{equation}
where $\EX_{\omega}[C_{\omega}]$ represents the same regularization as the previous method sampling $\omega \sim Q_{\theta}(\omega)$, and $K$ is a hyperparameter that adjusts the trade-off between the end-to-end loss and the regularization term. Note that $\boldsymbol{z}^*(h_{\omega}, \boldsymbol{x_i}^t, M^t)$ represents a Stochastic OP solution with $d_{z_d} + M^t*d_{u}$ decision variables. While this approach does increase the training time since the Stochastic OP needs to be solved during the training process, our hypothesis is that the sampling size $M^t$ does not need to be high. This is because the final task solution is not dependent on the accurate reconstruction of the actual data distribution, but on a learned latent distribution.

Gradient descent is used during the training process, and computing the gradients of the right-hand side of Equation \eqref{eqn:ENDLOSS} with respect to $\theta$ requires to solve the challenge of computing the chain $\frac{\partial f}{\partial z^*} \frac{\partial z^*}{\partial h_{\omega}} \frac{\partial h_{\omega}}{\partial \theta}$. Specifically, the partial $\frac{\partial z^*}{\partial h_{\omega}}$ is computed through the Stochastic Programming block, i.e., Equation \ref{eqn:PaOMonteCarlo}, and the argmin differentiation can be complicated because the gradients have to be computed through an OP solver. To overcome this, we leverage specialized methods \cite{amos2017optnet,agrawal2019differentiable} to perform KKT differentiation.

The learning process yields BNN parameters $\theta^*$, and then we solve Equation \eqref{eqn:PaOMonteCarlo} during decision inference given a new input $\boldsymbol{x}$.

\section{Evaluation}
\label{evaluation}

In decision theory, the quality of a decision is often evaluated using the regret metric \cite{bell1982regret,birge2011introduction,Demirovic2019AnII}. The average regret, $R$, for a dataset $\mathcal{D} =\{ (\boldsymbol{x_i}, \boldsymbol{y_i} ) \}_{i=1}^{N}$ with a trained BNN model $h_{\omega}$ is given by $R = \frac{1}{N}\sum_{i=1}^N f(\boldsymbol{z}^*({\boldsymbol{x_i}, h_{\omega}}), \boldsymbol{y}_i) - f(\boldsymbol{z}^*({\boldsymbol{y}_i}), \boldsymbol{y}_i)$. However, this metric may not accurately reflect model performance on noisy data. To address this, we also calculate a free-aleatoric version of the regret ($FR$), or the expected regret, defined when the data distribution $P(\boldsymbol{y} \mid \boldsymbol{x})$ is known, crucial for a proof of concept in synthetic problems: 
\[
FR = \frac{1}{N}\sum_{i=1}^N f(\boldsymbol{z}^*({\boldsymbol{x_i}, h_{\omega}}), \boldsymbol{y}_i) - f(\boldsymbol{z}^*({\boldsymbol{y}^{dist}_i}), \boldsymbol{y}_i),
\]
where $\boldsymbol{z}^*({\boldsymbol{y}_i^{dist}})$ is the argmin over the actual conditional distribution, computed as
\[
\boldsymbol{z}^*({\boldsymbol{y}_i^{dist}}) := \argmin_{\boldsymbol{z}} \mathbb{E}_{{{\boldsymbol{y}_i^{dist}} \sim P(\boldsymbol{y}_i \mid \boldsymbol{x_i})}} [f(\boldsymbol{z}, \boldsymbol{y}_i^{dist})] \quad \text{subject to } \boldsymbol{z} \in {S}.
\]
An effective model has $FR << R$, indicating the remaining regret is due to data noise. These metrics are leveraged to evaluate methods on synthetic datasets through Monte Carlo simulations in our experiments, while in real-datasets we compute only $R$.

\section{Experiments}
\label{experiments}

In this section, we design three data-driven OPs for evaluation, detail the chosen baselines, and present the experimental results. Further implementation details can be found in Appendix D.

\subsection{Classical Newsvendor Problem}
\label{op1}

The classical newsvendor (NV) problem is defined as finding the optimal order quantity $z^*$ that minimizes the cost function $z^*({y}) = \argmin_{z} c_s (y-z)_+ + c_e (z-y)_+$, subject to $z\geq0$, where $(u)_+ = \max(u,0)$ for demand $y \in \mathbb{R}$ and $c_s$, $c_e$ denote shortage and excess unit costs respectively. The goal is to estimate demand $y$ to minimize costs; ideally, $z^*(y) = y$ leads to zero cost. When demand follows a distribution, the optimal $z^*$ aligns with the quantile $\frac{c_s}{c_s+c_e}$ of this distribution, offering a closed-form solution for minimizing expected costs \cite{ban2019big}.


\paragraph{Data and OP parameters.} We generate datasets of $(x, y)$ pairs with non-linear relationships: 1800 for training, 1200 for validation, and 1200 for testing, where $x, y \in \mathbb{R}$. The data is used in two Newsvendor (NV) experiments, NV1 and NV2, introducing input-dependent Gaussian noise and Multimodal Gaussian noise, respectively, to simulate heteroscedastic uncertainty. Additionally, varying densities in the input space are used to simulate epistemic uncertainty. 
We set $c_s = 100$ and $c_e = 900$ to emphasize the cost imbalance. Appendix A.2 discusses that equalizing $c_s$ and $c_e$ negates the advantage of uncertainty propagation.

\subsection{Quadratic Programming Newsvendor}
\label{op2}
We now consider a constrained and quadratic version of the Newsvendor problem (NVQP) with multiple items defined by the following equation:
\begin{equation}
\label{eqn:constrainedNV}
\begin{aligned}
\boldsymbol{z}^*({\boldsymbol{y}}) = &\argmin_{\boldsymbol{z}} \boldsymbol{z}^\intercal \boldsymbol{Q} \boldsymbol{z} 
+ \boldsymbol{c}^\intercal \boldsymbol{z} + (\boldsymbol{y}-\boldsymbol{z})_+^\intercal \boldsymbol{Q_s} (\boldsymbol{y}-\boldsymbol{z})_+ 
+ \boldsymbol{c_s}^\intercal (\boldsymbol{y}-\boldsymbol{z})_+ \\
&+ (\boldsymbol{z}-\boldsymbol{y})_+^\intercal \boldsymbol{Q_e} (\boldsymbol{z}-\boldsymbol{y})_+
+ \boldsymbol{c_e}^\intercal (\boldsymbol{z}-\boldsymbol{y})_+ \quad \text{subject to } \boldsymbol{z} \succeq0 \text{ and } \boldsymbol{p}^\intercal \boldsymbol{z} \leq B\\
\end{aligned}
\end{equation}
where $\boldsymbol{y}$ is the unknown demand, $\boldsymbol{Q}, \boldsymbol{Q_s}, \boldsymbol{Q_e} \in \mathbb{R}^{d_z \times d_z}$, and $\boldsymbol{c}, \boldsymbol{c_s}, \boldsymbol{c_e} \in \mathbb{R}^{d_z}$ are quadratic and linear deterministic parameters in the cost function regarding fixed, shortage and excess costs of each item, and $\boldsymbol{p} \in \mathbb{R}^{d_z}$ and $B \in \mathbb{R}$ are deterministic parameters in the inequality constraint regarding the unit price of items and the total budget. In this problem, $d_y=d_z$. With a few mathematical steps detailed in Appendix B, this problem is transformed into a standard Quadratic Programming formulation (i.e., $\argmin_{\boldsymbol{v}} \frac{1}{2}\boldsymbol{v}^\intercal \boldsymbol{H} \boldsymbol{v} + \boldsymbol{k}^\intercal \boldsymbol{v}$ s.t. $\boldsymbol{A}\boldsymbol{v}\preceq \boldsymbol{b}$).

\paragraph{Data and OP parameters.}
We generate 4000 / 2000 / 2000 training / validation / test pair samples $(\boldsymbol{x}, \boldsymbol{y})$, where $\boldsymbol{x} \in \mathbb{R}^4$ and $\boldsymbol{y} \in \mathbb{R}^6$ (i.e., $d_x=4$ and $d_y=6$), with a nonlinear and noisy relationship between those variables, similar to the previous experiment. From each data sample, we seek to find $\boldsymbol{z}^*(\boldsymbol{x}) \in \mathbb{R}^6$ (i.e., $d_z=6$). The noise of \mbox{$\boldsymbol{y} \mid \boldsymbol{x}$} is generated by mixing different class of distributions across the outputs.

\subsection{Portfolio Conditional Loss Minimization}
\label{op3}

Drawing from Conditional Value at Risk formulation \cite{rockafellar2000optimization}, we address a Portfolio Optimization Problem (POP) aiming to minimize potential losses exceeding a threshold (here, zero) by resource allocation $\boldsymbol{z}$ across assets, given uncertain asset performance $\boldsymbol{y}$. The optimization is framed as $\boldsymbol{z}^*({\boldsymbol{y}}) = \argmin_{z}( \max{-\boldsymbol{y}^\intercal \boldsymbol{z}, 0})$, constrained by $\boldsymbol{z} \succeq0$ and a minimum expected return $\boldsymbol{p}^\intercal \boldsymbol{z} \geq R$, with $\boldsymbol{p}$ representing historical average returns. Appendix C explains transforming this into a Linear Programming problem, then approximating it as Quadratic Programming for compatibility with a quadratic solver, including a regularization term for solution refinement based on \cite{wilder2018melding}.

\paragraph{Data and OP parameters.} In this experiment, we use both synthetic (POP) and real datasets (POP2). As a synthetic dataset, we generate 1500 / 900 / 1500 training / validation / test pair samples $(\boldsymbol{x_i}, \boldsymbol{y_i})$, where $\boldsymbol{x_i} \in \mathbb{R}^3$ and $\boldsymbol{y_i} \in \mathbb{R}^{15}$ (i.e., $d_x=3$ and $d_y=15$), with a nonlinear relationship similar to the previous experiments. We also vary the number of the training dataset and the number of assets ($d_y$) for further analysis in the results section. The real dataset \cite{hoseinzade2019cnnpred} includes daily data from 2010 to 2017 on major US stock indexes and features such as technical indicators, futures, commodity prices, global market indices, major US company prices, and treasury bill rates.

\subsection{Methods and Baselines}
\label{baselines}

\paragraph{Predictors.} We implemented both methods BNN \emph{Decoupled} and BNN \emph{Combined} (we denote in this section as D-BNN and C-BNN, respectively) with a fully connected architecture. For the NV problem, we used three hidden layers with (128, 64, 64) neurons. For the NVQP and POP, we used three hidden layers with (512, 128, 128) neurons. We consider the respective standard neural networks ANN \emph{Decoupled} and ANN \emph{Combined} as baselines (we denote in this section as D-ANN and C-ANN, respectively), with the same number of hidden layers and neurons but without the uncertainty modeling. For the C-ANN baseline, our implementation is based on the \cite{donti2019mloc} idea with a single output (instead of a fixed number of categories) for the NVQP, and based on the \cite{wilder2018melding} (Linear Programming version with a quadratic additional term) for the POP experiment. We also implement Gaussian process (GP) as a decoupled baseline since they are commonly used for predictions' uncertainty. We considered the GP with radial basis function kernel with the white noise addition; this combination of kernels provided better results than other kernels and could model epistemic and homoscedastic aleatoric uncertainty. 

\paragraph{OP solvers.} In the NV problem, $z^*$ and its gradients are computed within a closed-form solution, so no specific mathematical solver is needed. In the NVQP and in the POP problems, $\boldsymbol{z}^*$ and the KKT differentiation were computed using the qpth library \cite{amos2017optnet}, which leverages the cvxpy quadratic programming solver.

\subsection{Results}
\label{results}

\paragraph{Main results.} 
Table \ref{table-main-results} shows the results by running it five times varying the seed data generation and computing the average and standard deviation values. The table is divided into the presented experiments. It shows that the D-ANN method has the highest $R$ and $FR$ (worst result) for the experiments. Both $R$ and $FR$ decrease when modeling uncertainty with the GPs and BNN in a \emph{Decoupled} fashion. Although the C-ANN is able to achieve reasonable results, we observed that it can sometimes converge to a bad local minima, resulting in a stagnation of the learning process, as it happened for the POP experiment. Finally, the C-BNN outperformed the other methods, but only with a small advantage compared to the D-BNN (in most cases). Indeed, we observed that the BNN sampling size that we use to approximate the expectation operations in training and inference ($M^t$ and $M$) plays an important role in the results. Therefore, we investigate important differences between the two versions of the presented BNNs in the following analysis.

\begin{table*}
	\centering
	\caption{Mean (and std between brackets) of $R$ and $FR$ for all the experiments. Some results in this table were scaled to be represented as an integer.}
	\begin{tabular}{lrrrrrrrrrrrrrr}
		\toprule
		& \vline&
		\multicolumn{2}{c}{Exp: NV1}&
		\vline&
		\multicolumn{2}{c}{Exp: NV2}&
		\vline&
		\multicolumn{2}{c}{Exp: NVQP}&
		\vline&
        \multicolumn{2}{c}{Exp: POP}&
        \vline&
        \multicolumn{1}{c}{Exp: POP2}        
		\\
		\textbf{Method}& \vline&
		\textbf{R} &
		\textbf{FR}& 
		\vline&
		\textbf{R} &
		\textbf{FR}&
		\vline&
		\textbf{R} &
		\textbf{FR}&
		\vline&
		\textbf{R} &
		\textbf{FR}	&
		\vline&
		\textbf{R}				
		\\
		\midrule 
		D - ANN &\vline &958 &531 (70) &\vline 
		&950 &589 (67) &\vline
		&1456 &419 (12) &\vline
		&2137 &1938 (79) &\vline
		&1228 (283)
		\\
		
		D - GP &\vline &617 &191 (16) &\vline 
		&583 &222 (20) &\vline
		&1337 &300 (15) &\vline
		&274 &75 (13) &\vline
		&\textbf{852 (31)}
		\\ 
		
		D - BNN &\vline &\textbf{460} &\textbf{33 (7)} &\vline 
		&\textbf{421} &\textbf{60 (8)} &\vline 
		&\textbf{1252} &\textbf{214 (11)} &\vline
		&\textbf{246} &\textbf{47 (15)} &\vline
		&944 (83)
		\\
		
		\midrule
		
		C - ANN &\vline &\textbf{461} &\textbf{33 (12)} &\vline 
		&\textbf{410} &\textbf{49 (22)} &\vline 
		&1263 &226 (30) &\vline
		&2138 &1939 (84) &\vline
		&1022 (61)
		\\
		
		C - BNN &\vline &\textbf{457} &\textbf{30 (3)} &\vline 
		&\textbf{400} &\textbf{39 (11)} &\vline 
		&\textbf{1242} &\textbf{204 (7)} &\vline
		&\textbf{245} &\textbf{46 (36)} &\vline
		&\textbf{721 (95)}
		\\
		\bottomrule
	\end{tabular}
	
	\label{table-main-results}
\end{table*}

\begin{table}[]	
	\centering
	\caption{The table shows the $FR$ average (out of five runs) result values for the NVQP experiment for different sampling sizes $(M^t, M)$.} 
	\begin{tabular}{|l|l|l|l|l|l|l|l|}
		\hline
		\textbf{Method} / ($\boldsymbol{M^t}$, $\boldsymbol{M}$) & \textbf{(4, 8)} & \textbf{(8, 8)} & \textbf{(8, 16)} & \textbf{(16, 16)} & \textbf{(16, 32)} & \textbf{(16, 64)} \\ \hline
		\textbf{FR (C-BNN)} & 259 & 234 & 221 & 217 & 209 & 204 \\
		\textbf{FR (D-BNN)} & 354 & 331 & 262 & 259 & 231 & 214 \\
		\hline
	\end{tabular}
	\label{table-exp-BNN-M}
\end{table}

\begin{figure}[htp] 
	\centering
	\subfloat[Values of $FR$ decrease by fixing $M^t = 32$ and increasing $M$.]{%
		\includegraphics[width=0.4\linewidth]{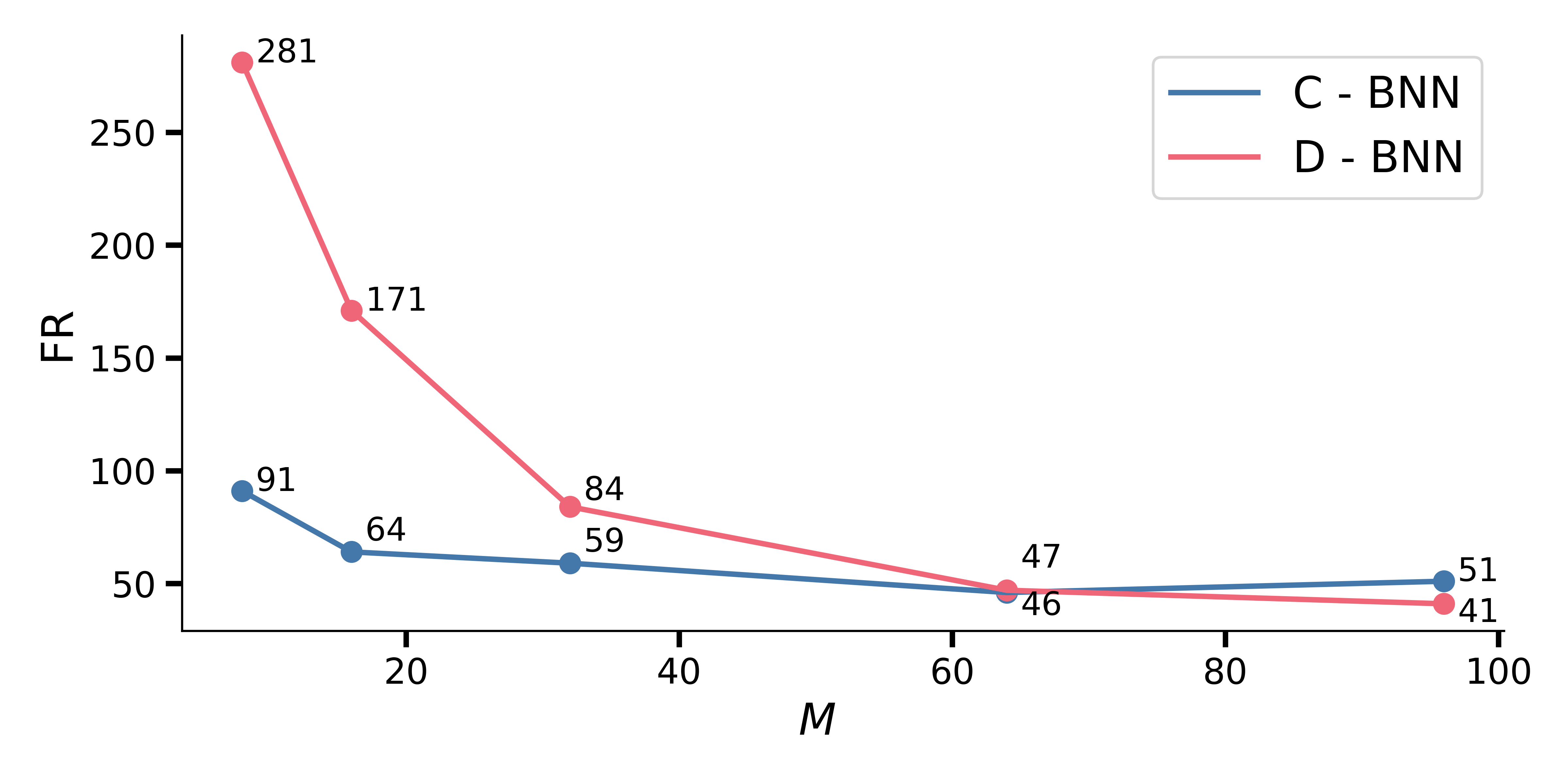}
		\label{fig:mopt}
	}%
	\hfill%
	\subfloat[$FR$ decreases by increasing the amount of training data available.]{%
		\includegraphics[width=0.4\linewidth]{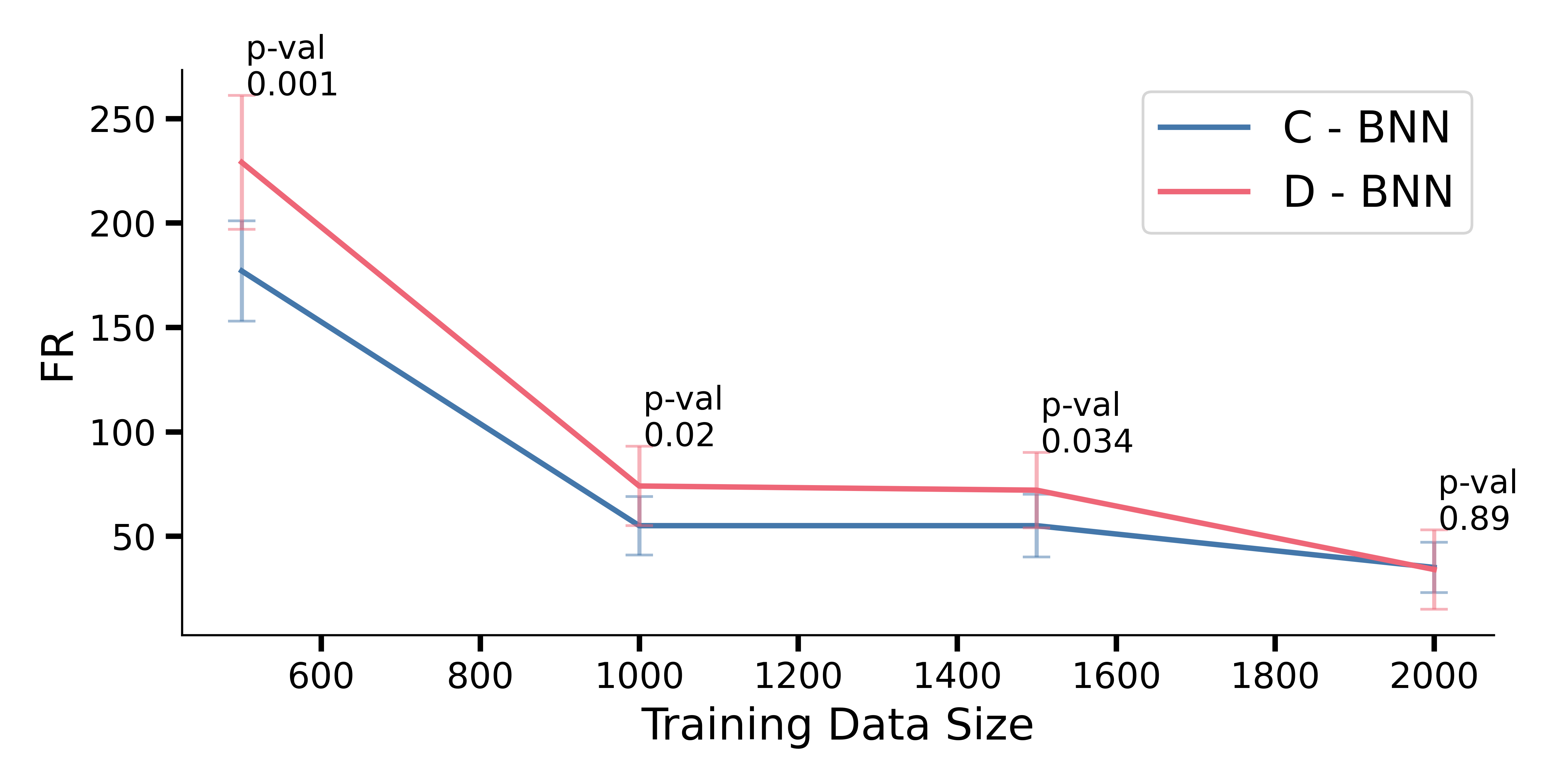}
		\label{fig:trainsize}
	}%
	\caption{Variation of $FR$ with sampling size and training data.}
\end{figure}

\paragraph{Varying the sampling size of BNNs.} 
To provide the main results for the NVQP experiment, the pair $(M^t, M)$ was limited to $(16, 64)$ for both D-BNN and C-BNN. In the POP experiment, we have limited both BNNs to $(M^t, M) = (32, 64)$. Initially, the idea is that these values should be as large as possible in order to approximate the expectation operations of Equations \ref{eqn:PaOMonteCarlo} and \ref{eqn:ENDLOSS}, but increasing the number of samples can lead to solving an OP with more decision variables, as detailed in the methods. In Table \ref{table-exp-BNN-M}, we show that, as expected, the values of $FR$ decrease (better) by increasing both $M^t$ and $M$ for the experiments NV1 and NVQP. The same is valid for increasing only $M$ while fixing $M^t$, as shown in Figure \ref{fig:mopt} for the POP experiment. It is observed that the C-BNN requires less sampling to converge to small values of $FR$. 

\paragraph{Varying the training dataset size.}

Fixing the sampling size to ($M^t = M = 32$) and for $d_y = d_z = 10$, we have also analyzed how the quality of the decisions varies with the increase in training data availability in the POP experiment. Figure \ref{fig:trainsize} shows that in a scenario with less training data, the C-BNN has significantly outperformed the D-BNN. The difference becomes insignificant as we increase the data size in the training set.

\begin{figure}
	\centering
	\includegraphics[width=0.8\linewidth]{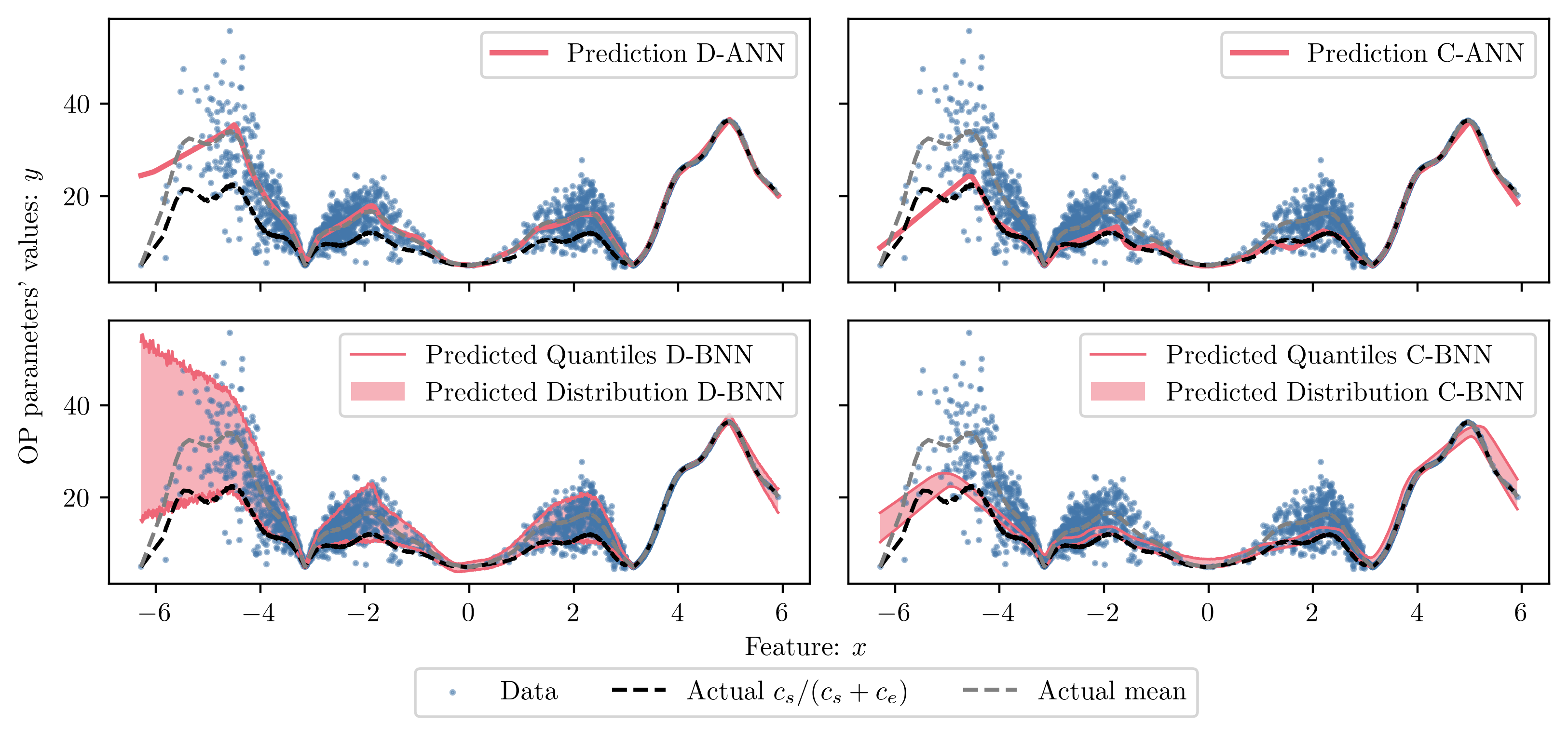}
	\caption{For the NV1 OP, the C-ANN, D-BNN, and D-BNN methods achieve good decisions with different strategies for the OP parameters predictions.} \label{fig:preds}
\end{figure}

\paragraph{Interpreting predicted OP parameters.}
Building upon the simplicity of our NV1 experiment, Figure \ref{fig:preds} illustrates the contrasting behaviors of different methods in predicting OP parameters. The D-ANN method successfully predicts the average of the unknown parameters' distribution (upper-left graph) but fails in the downstream task due to data noise (Table \ref{table-main-results}). The D-BNN method captures the uncertainty of the OP parameters, improving final task performance by aligning the $\frac{c_s}{c_s+c_e}$  quantile of the predicted distribution with the actual one (bottom-left graph) if enough sampling size is chosen. Conversely, the C-ANN method, shown in the upper-right graph, predicts the $\frac{c_s}{c_s+c_e}$ quantile rather than the mean and performs well in the final task, but exhibits diminished performance in more complex OPs due to a lack of uncertainty modeling. Lastly, the C-BNN method (bottom-right graph) learns a distribution that minimize the OP cost; even though the overall predicted distribution may not align closely with the actual one, the congruence of the $\frac{c_s}{c_s+c_e}$ quantiles illustrates its ability to focus on the most relevant aspect of the distribution. In both D-BNN and C-BNN graphs, the lower predicted quantile represents the $\frac{c_s}{c_s+c_e}$ quantile of the predicted distribution.

\section{Related Work}

A common method of data-driven decision-making, known as ``predict-then-optimize", is first to predict the unknown parameters of the OP and then using a solver to yield optimal decisions. This method has been criticized for propagating prediction errors to the optimization problem (OP) block \cite{elmacht2017spo}. This led to the development of ``Smart predict, then optimize," which incorporates OP information into the learning process with a surrogate end-to-end loss function, though it focuses only on linear objectives and linear predictors.

Following the introduction of KKT differentiation \cite{amos2017optnet}, subsequent works have combined ANNs with Quadratic Programming solvers \cite{donti2019mloc}, increasing predictive complexity compared linear models. These methods laid the groundwork for linear programming with neural networks \cite{wilder2018melding} and relaxation techniques for discrete OPs \cite{ferber2019mipaal,lahoud2024datasp}.

Like our approach, \cite{donti2019mloc,kong2022end} aimed to minimize the expected objective function value stochastically. While \cite{donti2019mloc} either lacked proper probabilistic modeling and manually discretized the target variables (OP parameters) before solving the OP or relied on analytical expectations, \cite{kong2022end} is an \emph{approximation} of a combined approach using energy-based models. Our method uniquely employs BNNs to model distributions within an task-based loss, enhancing decision quality through proper distribution modeling.

Other works have considered modeling predictions' distribution with BNNs \cite{pearce2020uncertainty,kendall2017uncertainties,blundell2015weight} but without focusing on solving stochastic or constrained OPs. Our work differs by adapting BNNs within a predictor-optimizer framework to improve data-driven decision quality.

\section{Conclusion}

This paper presented a framework for solving uncertain optimization problems (OPs) using input-output training data by predicting unknown parameters probabilistically and applying a Stochastic Programming technique for near-optimal decisions. 
We used BNNs to treat predictions as distributions and presented two ways of learning their weights. 

The proposed \emph{Decoupled} BNN models aleatoric and epistemic uncertainty, leveraging Variational Inference techniques to predict the OP parameters' distribution. It revives decoupled approaches value by providing good decision results. Also, it can be used in OPs where differentiation over the argmin operator is impossible or time-consuming. The proposed \emph{Combined} BNN learning approach, on the other hand, focused on directly minimizing the expected cost of the OP in an end-to-end fashion through a differentiable solver. Although its training process is more time consuming, we showed that it considerably outperforms state-of-the-art combined approaches in non-trivial OPs. It also outperforms the \emph{Decoupled} BNN mainly in scenarios where there is a limitation of sampling size and training data size.

\subsubsection*{Acknowledgement}
This work has been supported by the Industrial Graduate School Collaborative AI \& Robotics funded by the Swedish Knowledge Foundation Dnr:20190128, and the Knut and Alice Wallenberg Foundation through Wallenberg AI, Autonomous Systems and Software Program (WASP).

This preprint has not undergone peer review or any post-submission correction. The Version of Record of this contribution is published in Lecture Notes in Computer Science (LNCS 15016).

\section{Appendices}

	\subsection{Appendix A. Limitations of Uncertainty Propagation}
	\label{appendix:limitations}
	
	This paper focuses on minimizing $\argmin_{\boldsymbol{z}} \EX_{\hat{\boldsymbol{y}}}f(\boldsymbol{z}, \hat{\boldsymbol{y}})$. This problem simplifies to $\argmin_{\boldsymbol{z}} f(\boldsymbol{z}, \EX_{\hat{\boldsymbol{y}}}\hat{\boldsymbol{y}})$ when substituting the objective function's expected value with the expected value of predictions, but this simplification is only applicable in certain conditions. If these conditions are met, we recommend solving the argmin by directly calculating the expected value of predictions (Decoupled).
	
	\subsubsection{Appendix A.1. Linear objective functions with respect to the unknown variable}
	\label{appendix:linearity}
	If $f(\boldsymbol{z}, \boldsymbol{y})$ is linear with respect to $\boldsymbol{y}$, then $\EX_{\hat{\boldsymbol{y}}}f(\boldsymbol{z}, \hat{\boldsymbol{y}}) = f(\boldsymbol{z}, \EX_{\hat{\boldsymbol{y}}}\hat{\boldsymbol{y}})$. Applying the argmin with respect to $\boldsymbol{z}$ on both sides we have $\argmin_{\boldsymbol{z}} \EX_{\hat{\boldsymbol{y}}}f(\boldsymbol{z}, \hat{\boldsymbol{y}}) = \argmin_{\boldsymbol{z}} f(\boldsymbol{z}, \EX_{\hat{\boldsymbol{y}}}\hat{\boldsymbol{y}}).$
	
	\subsubsection{Appendix A.2. Balanced Newsvendor Problem}
	\label{appendix:imbalance}
	When $c_s = c_e$ in the NV problem, the optimal order quantity $\argmin_{z} \EX_{\hat{y}}f(z, \hat{y})$ corresponds to the median of $\hat{y}$'s distribution, given by the $\frac{c_s}{c_s+c_e} = 0.5$ quantile. If $\hat{y}$'s distribution is Gaussian, this median equals the mean, simplifying the argmin to the mean of $\hat{y}$. This observation extends to both Gaussian models and the Quadratic Programming (QP) approach in the Newsvendor Problem, highlighting that propagating uncertainty becomes more beneficial as the imbalance between $c_s$ and $c_e$ increases.
	
	\subsection{Appendix B. Newsvendor Problem as Quadratic Programming}
	\label{appendix:NVQP}		
	Following \cite{donti2019mloc} and \cite{rudin2018newsvendor}, we reformulate Equation 8 by introducing new decision variables $\boldsymbol{z_s} = \boldsymbol{y}-\boldsymbol{z}$ and $\boldsymbol{z_e} = \boldsymbol{z}-\boldsymbol{y}$, with added constraints to align with the original problem's bounds. This leads to a QP formulation: $\argmin_{\boldsymbol{v}} \frac{1}{2}\boldsymbol{v}^\intercal \boldsymbol{H} \boldsymbol{v} + \boldsymbol{k}^\intercal \boldsymbol{v}$ subject to $\boldsymbol{A}\boldsymbol{v}\preceq \boldsymbol{b}$, where $\boldsymbol{H} = 2\text{diag}[\boldsymbol{Q}, \boldsymbol{Q_s}, \boldsymbol{Q_e}]$, $\boldsymbol{v} = [\boldsymbol{z}, \boldsymbol{z_s}, \boldsymbol{z_e}]$, $\boldsymbol{k} = [\boldsymbol{c}, \boldsymbol{c_s}, \boldsymbol{c_e}]$, $\boldsymbol{A} = [-\boldsymbol{I_{3d_z}}, [-\boldsymbol{I_{d_z}}, -\boldsymbol{I_{d_z}}, 0], [\boldsymbol{I_{d_z}}, 0, -\boldsymbol{I_{d_z}}], [\boldsymbol{p}, 0, 0]]^{\intercal}$, and $\boldsymbol{b}=[0, 0, 0, -\boldsymbol{y}, \boldsymbol{y}, B]$. $\boldsymbol{z}^*(\boldsymbol{y})$ is the primary variable of interest. Assuming $\boldsymbol{H}$ is positive-definite ensures convexity. The formulation's efficiency depends on the item count. It's initially suitable for single vector predictions $\boldsymbol{y}$, but we propose a Stochastic Programming method for generalization to multiple predictions.

	\subsubsection{Appendix B.1. Newsvendor Problem as Stochastic Quadratic Programming }
	
	When propagating the uncertainty of $\boldsymbol{y}$ in a Monte Carlo fashion with $M$ samples, the formulation above becomes as $\argmin_{\boldsymbol{v}} \frac{1}{2}\boldsymbol{v}^\intercal \boldsymbol{H} \boldsymbol{v} + \boldsymbol{k}^\intercal \boldsymbol{v}$ s.t. $\boldsymbol{A}\boldsymbol{v}\preceq \boldsymbol{b}$ where $H = 2diag([\boldsymbol{Q} \quad \frac{\boldsymbol{Q_s}}{M} \quad ... \quad \frac{\boldsymbol{Q_s}}{M} \quad \frac{\boldsymbol{Q_e}}{M} \quad ... \quad \frac{\boldsymbol{Q_e}}{M}])$; $\boldsymbol{k} = [\boldsymbol{c} \quad \frac{\boldsymbol{c_s}}{M} \quad ... \quad \frac{\boldsymbol{c_s}}{M} \quad \frac{\boldsymbol{c_e}}{M} \quad ... \quad \frac{\boldsymbol{c_e}}{M}]$; $	\boldsymbol{v} = [\boldsymbol{z} \quad \boldsymbol{z_s}^{(1)} \quad ... \quad \boldsymbol{z_s}^{(M)} \quad \boldsymbol{z_e}^{(1)} \quad ... \quad \boldsymbol{z_e}^{(M)}]$; $\boldsymbol{A} = [-\boldsymbol{I_F}, 
	[-\boldsymbol{I_{B1}}, -\boldsymbol{I_{BB}}, \boldsymbol{0_{BB}}], 
	[\boldsymbol{I_{B1}} ,\boldsymbol{0_{BB}} , -\boldsymbol{I_{BB}}], 
	[\boldsymbol{p} ,0 , 0]]^\intercal$; 
	and \\ $\boldsymbol{b} = [\boldsymbol{0_{F}}, -\boldsymbol{y}^{(1)}, ..., -\boldsymbol{y}^{(M)}, \boldsymbol{y}^{(1)},..., \boldsymbol{y}^{(M)}, B]$. 
	Where $\boldsymbol{I_F} = \boldsymbol{I_{d_z + 2Md_z}}$, 
	$\boldsymbol{0_F} = \boldsymbol{0_{d_z + 2Md_z}}$ (1D vector), 
	$\boldsymbol{I_{BB}} = \boldsymbol{I_{Md_z}}$, 
	$\boldsymbol{0_{BB}} = \boldsymbol{0_{Md_z}}$, 
	$\boldsymbol{I_{B1}} = \boldsymbol{I_{d_z}}$ repeated for $M$ rows. 
	This is a generalization of the quadratic newsvendor experiment proposed in [Donti et al., 2017]. Note that $v \in \mathbb{R}^{d_z+2Md_z}$, $H \in \mathbb{R}^{d_z+2Md_z \times d_z+2Md_z}$, $k \in \mathbb{R}^{d_z+2Md_z}$, $A \in \mathbb{R}^{d_z + 4Md_z+1 \times d_z+2Md_z}$ and $b \in \mathbb{R}^{d_z + 4Md_z+1}$. Therefore, both the number of items and prediction sampling size play an important and approximately equal role on the time to solve each instance of the OP. In practice, the complexity of the QP problem depends on the decision variable dimension, which is $d_z+2Md_z$.

	\subsection{Appendix C. Portfolio Risk Minimization as a Linear Programming}
	With the same strategy as in Appendix \ref{appendix:NVQP}, we use the auxiliary variable  $\boldsymbol{u} = \max\{-\boldsymbol{y}^\intercal \boldsymbol{z}, 0\}$ to rewrite the POP formulation from the main text to 
	\begin{equation}
	\label{eqn:portfRiskLP}
	\begin{aligned}
	&(\boldsymbol{z}^*, \boldsymbol{u}^*)({\boldsymbol{y}}) = \argmin_{\boldsymbol{z}, \boldsymbol{u}} \boldsymbol{0}^\intercal \boldsymbol{z} + \boldsymbol{u} \quad \text{s.t. } -\boldsymbol{[z, u]}\preceq0, 
	-\boldsymbol{y}^\intercal \boldsymbol{z} \preceq \boldsymbol{u},
	-\boldsymbol{p}^\intercal \boldsymbol{z} \leq -R.
	\end{aligned}
	\end{equation}
	Note that the zero constant in the objective function is only to reinforce that $\boldsymbol{z}$ is also part of the set of decision variables.
	
	\subsubsection{Appendix C.1. Portfolio Risk Minimization as a Stochastic Linear Programming}
	
	By giving a set of samples $\boldsymbol{y}^{(j)}$ as input, as suggested in [Rockafellar et al., 2000], the equation above can be rewritten in a stochastic programming fashion as 
	
	\begin{equation}
	\label{eqn:portfRiskSLP}
	\begin{aligned}
	&(\boldsymbol{z}^*({\boldsymbol{y}}), \boldsymbol{u}^*({\boldsymbol{y}}) ) = \argmin_{\boldsymbol{z}, \boldsymbol{u}} \boldsymbol{0}^\intercal \boldsymbol{z} + \frac{1}{M} \sum_{j=1}^{M} \boldsymbol{u}^{(j)}\\ 
	&\text{s.t. } -\boldsymbol{z}\preceq0, -\boldsymbol{u}^{(j)} \preceq \boldsymbol{0}, -\boldsymbol{y}^{\intercal (j)} \boldsymbol{z} -\boldsymbol{u}^{(j)} \preceq \boldsymbol{0} \quad \forall j \in 1..M,
	&-\boldsymbol{p}^\intercal \boldsymbol{z} \leq -R.
	\end{aligned}
	\end{equation}
	For implementation purpose, we followed \cite{wilder2018melding} by adding a quadratic small term to linear programs in order to fit the OP into the Amos \& Kolter QP solver. 

	\subsubsection{Appendix D. Implementation details}
	\label{implem}
	Neural networks were implemented with Pytorch and the Adam optimizer, with learning rates of $0.0015$ for NV, $0.002$ for NVQP, and $0.001$ for POP experiments. The \emph{Decoupled} Bayesian Neural Network (BNN) had a learning rate of $0.0007$, whereas the \emph{Combined} BNN's rate ranged between $0.0004$ and $0.0007$. An exponential scheduler was used to adjust the learning rate by a factor of $0.99$. Hyperparameter $K$ balanced data loss and regularization, selected without optimization. Training occurred on Nvidia RTX 2080 GPUs, with models evaluated on the validation set before testing. Gaussian Process baselines, managed with Scikit-learn and a radial basis function kernel, optimized the length scale and white noise. For multi-output tasks (NVQP and POP), separate Gaussian processes for each output proved more effective. 

\bibliographystyle{splncs04}
\bibliography{mybibliography}
%





\end{document}